\def\year{2017}\relax
\documentclass[letterpaper]{article}
\usepackage{aaai17}
\usepackage{times}
\usepackage{helvet}
\usepackage{url}
\usepackage{courier}
\usepackage{amsfonts}
\usepackage{amsmath}
\usepackage{color}
\usepackage{graphicx}
\usepackage{tabularx}
 \usepackage{multirow}
\usepackage{paralist}
\begin{document}
\title{Improving Multi-Document Summarization via Text Classification }
\author{
Ziqiang Cao$^{1,2}$ ~~ ~~ Wenjie Li$^{1,2}$ ~~ ~~ Sujian Li$^3$ ~~ ~~ Furu Wei$^4$\\
$^1$Department of Computing, The Hong Kong Polytechnic University, Hong Kong\\
$^2$Hong Kong Polytechnic University Shenzhen Research Institute, China\\
$^3$Key Laboratory of Computational Linguistics, Peking University, MOE, China \\
$^4$Microsoft Research, Beijing, China\\
{\tt \{cszqcao, cswjli\}@comp.polyu.edu.hk  } \\
{\tt lisujian@pku.edu.cn} \\
{\tt fuwei@microsoft.com} \\
}
\maketitle
\begin{abstract}
	\begin{quote}
		Developed so far, multi-document summarization has reached its bottleneck due to the lack of sufficient training data and diverse categories of documents.
		Text classification just makes up for these deficiencies. 
		In this paper, we propose a novel summarization system called TCSum, which leverages plentiful text classification data to improve the performance of multi-document summarization. 
		TCSum projects documents onto distributed representations which act as a bridge between text classification and summarization.
		It also utilizes the classification results to produce summaries of different styles.
		Extensive experiments on DUC generic multi-document summarization datasets show that, TCSum can achieve the state-of-the-art performance without using any hand-crafted features and has the capability to catch the variations of summary styles with respect to different text categories. 
		%Both summarization and text classification require models to understand the semantics of documents.
		%In addition, different text categories hold different summary styles.
		%Therefore, in this paper, we propose a novel summarization system called TCSum, which leverages plentiful text classification data to improve the performance of summarization.
		%TCSum projects documents onto distributed representations shared by the two tasks.
		%Meanwhile, it represents the summary in various ways according to the predicted document category.
		%We conduct extensive experiments on DUC generic summarization datasets.
		%Without using any hand-crafted features, TCSum achieves the state-of-the-art performance.
		%We also observe that TCSum indeed catches the variations of the summary styles among different text categories. 
	\end{quote}
\end{abstract}

\section{Introduction}
The increasing online information has necessitated the development of effective automatic multi-document summarization systems.
%So far, most summarization systems adopt the extractive mode~\cite{over2001introduction} which directly selects existing salient sentences to form the summary.
%Basically, there are two components in extractive summarization, i.e., sentence ranking and sentence selection.
%Sentence ranking, the core part of extractive summarization, measures the saliency of a sentence.
%Meanwhile, sentence selection chooses both salient and non-redundant sentences to form the summary.
Through long-term research, the learning-based summarization approaches have grown to become dominant in the literature. 
%However, the quality of a machine generated summary is still far from satisfactory. 
%A big bottleneck of supervised summarizers is the lack of human summaries used for training.
By far, a prominent issue that hinders the further improvement of supervised approaches is the lack of sufficient human summaries used for training~\cite{ziqiang2016tgsum}.
For instance, the widely-used DUC\footnote{\url{http://duc.nist.gov/}} generic multi-document summarization benchmark datasets contain less than 400 human reference summaries in total.
Writing summaries is an extremely labor-intensive and time-consuming process. 
Because of the limitation of training data, a learning-based summarization system is often forced to heavily rely on well-designed features. 
Simple models like Support Vector Regression can achieve the state-of-the-art performance with extensive linguistic and statistical features \cite{hong2014improving}.
%In certain extreme cases, unsupervised approaches (e.g., \cite{rioux2014fear}) even outperform supervised approaches.
To break through the bottleneck of insufficient summarization training data, taking advantage of other rich data sources might be a good idea worth considering.

Meanwhile, existing summarization approaches basically apply a uniform model to generate summaries for the documents in different text categories. 
However, according to what we observe, summary styles in different categories can vary to a large degree. 
Take the two common categories in DUC datasets, i.e., Natural Disaster and Biography as an example.
To summarize a natural disaster like a hurricane, people tend to present its moving path and the loss it brings.
%They focus on the power of the disaster and the loss it brings.
By contrast, a biography summary is expected to include the personal profile and the main contributions of the person.
Apparently, summaries should focus on different aspects of the topics which belong to the corresponding categories.
%The style problem may partly explain the experiment result of \cite{wan2015multi}.
When the document category is given, \cite{kedzie-mckeown-diaz:2015:ACL-IJCNLP}
 finds that the introduction of category-specific language models largely promotes the summarization performance.
The experiments of \cite{wan2015multi} also show that a summarization model with good overall performance still produces low-quality summaries in certain document sets.
The summary style issue previously mentioned may partly explain these phenomena and suggest a possible way to improve the summarization performance.

Compared with summarization, the text classification datasets are much richer.
Note that both summarization and text classification require models to understand the semantics of documents.
Better text representations learned by classification data can help to train more effective summarization models.
Moreover, if we know the category of a document, we will have a chance to explore more proper summary styles.
To this end, we propose a novel summarization system called TCSum, which leverages text classification data to improve the performance of summarization.
Since distributed representations of documents have demonstrated advantages in both summarization (e.g., \cite{kobayashi-noguchi-yatsuka:2015:EMNLP}) and text classification (e.g., \cite{lai2015recurrent}), TCSum projects all documents onto the distributed representations that are shared by the two tasks.
Then, for text classification, the document embeddings are followed by a classifier to learn their association with the categories. 
For summarization, the document embeddings are transformed to match the ``meaning'' of the reference summaries.
To make the transformed embeddings also hold the information of summary styles, we utilize the classification result and develop a category-specific transformation process.
Our model adopts the recent hot topic of neural network based transfer learning (e.g., from syntactic parsing to discourse parsing \cite{li2014recursive}).
It is also noted that our model is totally data-driven, i.e., all the abstract features are learned automatically.

%Inspiration: distributed representation for classification/summarization
%Our solution: Introduce text classification datasets to improve summarization performance.
%Document embeddings shared by the two tasks.
%For classification, add a softmax layer to learn the text class.
%For summarization, add a transform layer to learn the summary embedding.
%The transform matrix is generated according to the softmax layer result.
%This practice is consistent with the changes of summarization styles in different text classes.

We verify the effectiveness of TCSum on DUC generic summarization benchmark datasets.
%In our preliminary experiments, we ignore the Relevance Model and supplement the query information with surface features.
%This simplified approach has attained comparable performance against two popular summarization systems.
%Therefore, we expect the total framework of AttSum can work even better. 
TCSum is able to compete with state-of-the-art summarization systems which usually heavily depends on hand-crafted features.
We also observe that TCSum indeed catches the variations of the summary styles among different text categories.  

The contributions of this paper are listed as follows:
\begin{itemize}
	\item We leverage text classification datasets to learn better document representations for summarization.
	\item We explore the variations of summary styles with respect to different text categories.
	\item We develop a competitive summarization system which does not need any hand-crafted features. 
\end{itemize}

\section{Method}

%$s$ sentence.
%$D$ document.
%$S_D$ summary for document $D$.
%In extractive summarization, $S_D \subset D$.
%$t_D \in [1,|C|]$, the real text category for $D$.
Let $D$ denote a document which is composed of a set of sentences $\{ {s_i}|i \in [1,N]\} $.
For text classification, we use $C$ to stand for the entire set of categories.
We assume $D$ belongs to one of $C$, i.e., $c_D \in [1,|C|]$ where $c_D$ represents the actual category for of the document $D$.
The \textbf{text classification model} is trained to predict a category for $D$.
For supervised sentence ranking required by learning-based summarization, each sentence holds a saliency score, usually measured with respect to the human summaries (hereafter the reference summaries).
The \textbf{summarization model} is expected to learn how to rank sentences in accord with the actual sentence saliency.  

%Meanwhile, for sentence ranking in supervised extractive summarization, each sentence holds a saliency score $r^t_s$, usually measured by reference summaries\footnote{Here the superscript ``t'' stands for the real value, and we use the superscript ``p'' to indicate a predicted value.}.
%A summarizer is expected to ranks sentences in accord with the real sentence saliency.  

%a summarizer needs to select sentences to form a summary $S \subset D$ which can cover the main idea of the document.

\begin{figure}
	\centering
	\includegraphics[width=0.99\linewidth]{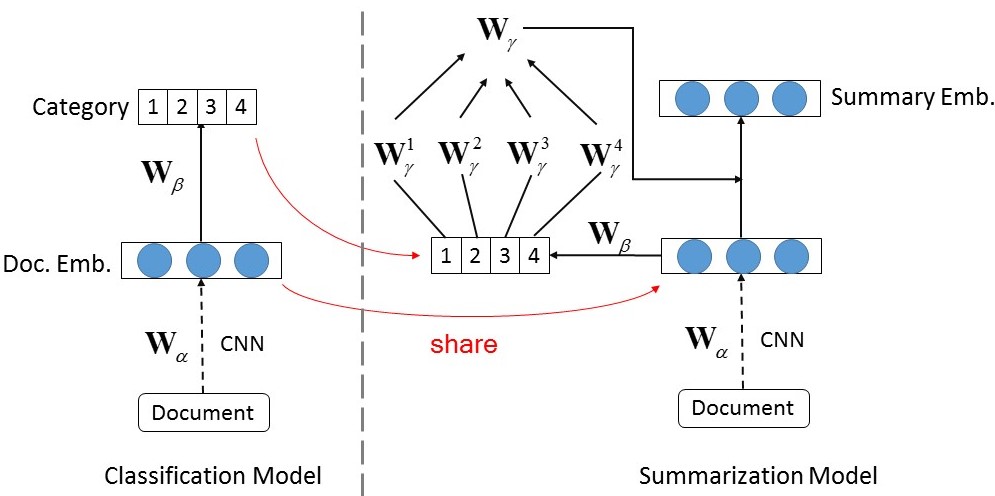}
	\caption{Overview of TCSum.}
	\label{fig:model}
\end{figure}

In this section, we describe how our summarization system, called TCSum, ranks the sentences with the help of text classification.
The overall framework of TCSum is illustrated in Fig.~\ref{fig:model}.
At first, a \textbf{text classification model} is trained using a convolutional neural network.
This model projects a document onto the distributed representation, and adds a softmax classifier to predict the category of the document. 
%Both the process to generate document embeddings and the predicted category will be utilized in summarization.
The \textbf{summarization model} shares the same projection process to generate document embeddings given that the semantic analysis and understanding of documents are essential for both classification and summarization.
%Considering that the text classification dataset is much larger than the summarization dataset, the generation process of document embeddings only learns from classification. 
Afterwards, it transforms the document embedding to the summary embedding and tries to maximize the match to the ``meaning'' of the reference summaries.
To make the transformed summary embedding sensitive to the different summary styles, TCSum learns category-specific transformation matrices according to the predicted categories.
%To make the transformed embedding also hold the information of summary styles, we develop a special layer to automatically build the transformation matrix according to the predicted category.
%Afterwards, we utilize the predict category to adaptively transform the document embedding into the summary embedding which matches the ``meaning'' of reference summaries.  
Finally, the sentences are ranked according to their saliency scores calculated based on the similarity between the sentence embedding and the summary embedding.
The rest of this section describes the details of our model.

%It firstly develops a text classification system with convolutional neural networks, which applies convolutional neural networks to project a document onto distributed representation
%It firstly project a document onto the distributed representation shared by both summarization and text classification. %since both of the two tasks need the understanding of the document meaning.
%Then, in the task of text classification, a softmax classifier is added to predict the category of the document.
%As for summarization, based on the predicted text category, TCSum adaptively transforms the document embedding into the summary embedding which matches the meaning of reference summaries. 

%Suppose ${s_i} = ({w_{i1}}, \cdots ,{w_{iK}})$ to be a sentence with word sequence $w_{i1}$ to $w_{iK}$.
%A document cluster $D$ is composed of sentences in it\footnote{We do not distinguish sentences from different documents for multi-document summarization.}, namely $D = \{ {s_i}\} $.

\subsection{Text Classification Model}
Convolutional Neural Networks (CNNs) can learn the abstract representations of N-grams effectively and tackle the sentences with variable lengths naturally.
Models using CNNs have achieved excellent performance both in text classification~\cite{lai2015recurrent} and summarization~\cite{yin2015optimizing}.
In this paper, we develop a simple CNN-based classification model.
Specifically, we use a CNN to project a sentence $s$ onto its distributed representation ${\mathbf{v}}(s) \in \mathbb{R}^m$, i.e., 
\begin{equation} \label{eq:sentenceEmbedding}
%{\bf{v}} (s) =\text{CNN}(s) 
s\xrightarrow{{{\text{CNN}}}}{\mathbf{v}}(s)
\end{equation}
%where ${{{\mathbf{W}}_\alpha }}$ is the weight matrix of CNNs.
A basic CNN contains a convolution operation on the top of word embeddings, which is followed by a pooling operation.
Let ${\bf{v}}(w_i) \in \mathbb{R}^k$ refer to the $k$-dimensional word embedding corresponding to the $i_{th}$ word in the sentence. 
Let ${\bf{v}}(w_i : w_{i+j})$ be the concatenation of word embeddings
$[{\bf{v}}(w_i),\cdots,{\bf{v}}(w_{i+j})]$. 
A convolution operation involves a filter ${\bf{W}}_\alpha \in \mathbb{R}^{m \times hk}$, which is applied to a window of $h$ words to produce the abstract features ${\bf{g}}_i^h \in \mathbb{R}^m$,
\begin{equation}\label{EQ:filter}
{\bf{g}}_i^h = f({\bf{W}}_ \alpha \times {\bf{v}}(w_i : w_{i+j})),
\end{equation}
where $f(\cdot)$ is a non-linear function and  the use of $tanh$ is the common practice. 
To make it simple, the bias term is left out.
This filter is applied to each possible window of words in the sentence to produce a feature map.
%\begin{equation}
%C^h = [c_1^h, \cdots , c_{N-h+1}^h]
%\end{equation}
%where $N$ is the length of the sentence. 
Subsequently, a pooling operation is applied over the feature map to obtain the final features $\hat{\bf{g}}^h \in \mathbb{R}^m$ of the filter.
Here we use the max-over-time pooling~\cite{Collobert2011}.
\begin{equation}
\hat{\bf{g}}^h=\max\{{\bf{g}}_1^h, {\bf{g}}_2^h,\cdots\}
\end{equation}
The primary purpose of this pooling is to capture the most important features in a feature map.
$\hat{\bf{g}}^h$ is the output of the CNN, i.e., the embedding of a sentence.

Then a document is represented by the average pooling of its sentence embeddings, just like~\cite{lai2015recurrent},
\begin{equation} \label{eq:documentEmbedding}
{\bf{v}}(D) = \frac{1}{{|D|}}\sum\nolimits_{s \in D} {{\bf{v}}(s)} 
\end{equation}
To learn the association between the document embedding and the categories, the document embedding is followed by a softmax classifier:
\begin{equation}\label{eq:classification}
{\bf{v}}_C(D) = {\rm{softmax}}({{\bf{W}}_\beta} \times {\bf{v}}(D)),
\end{equation}
where ${{\bf{W}}_\beta} \in \mathbb{R}^{|C| \times m}$ is the weight matrix, and ${\bf{v}}_C(D) \in \mathbb{R}^{|C|}$ is the predicted probability distribution over the categories.

%CNNs have shown promising progress in many NLP areas including summarization ~\cite{cao2015learning,yin2015optimizing}.
%Therefore, AttSum applies CNNs to learn semantic representations for sentences and queries.

%\cite{Collobert2011} first introduced CNN for a lot of word-related tasks including part-of-speech tagging, chunking, named entity recognition and semantic role labeling.
% Recently, CNN has been applied to model sentences~\cite{kalchbrenner2014convolutional} and achieved excellent results in semantic
%parsing~\cite{yih2014semantic}, search query retrieval~\cite{shen2014learning} as well as relation classification~\cite{zeng2014relation}.

\subsection{Summarization Model}
As previously mentioned, the summarization model in TCSum, shares the same convolution and pooling operations with the classification model when generating the document embedding ${\bf{v}}(D)$.
Then, TCSum transforms ${\bf{v}}(D)$ to match the ``meaning'' of the reference summary, i.e.,
\begin{equation}
{\bf{v}}_S(D) = {\rm{tanh}}({{\bf{W}}_\gamma} \times {\bf{v}}(D)),
\end{equation}
where ${\bf{v}}_S(D) \in \mathbb{R}^m$ is the transformed embedding called summary embedding, and ${\bf{W}}_\gamma \in \mathbb{R}^{m \times m}$ is the transformation matrix.
Note that we define the same dimension for both document and summary embeddings.
This setting simplifies the sentence ranking process, which is explained later.

We would also like the summary embedding to hold the information of summary styles. 
Inspired by the work of \cite{dong2014adaptive}, we develop the category-specific transformation matrix ${\bf{W}}_\gamma$ according to the predicted category.
We introduce $|C|$ sub-matrices $({\bf{W}}_\gamma ^1, \cdots ,{\bf{W}}_\lambda ^{|C|})$, with each directly corresponding to one text category.
Based on the predicted category derived from Eq.~\ref{eq:classification}, the transformation matrix ${\bf{W}}_\gamma$ is computed as the weighted sum of these sub-matrices.
\begin{equation}~\label{eq:adaptive}
{{\bf{W}}_\gamma } = \sum\nolimits_{i = 1}^{|C|} {{\bf{v}}_C^i(D)} {\bf{W}}_\gamma ^i
\end{equation}
In this way, ${{\bf{W}}_\gamma }$ is automatically biased to the sub-matrix of the predicted text category.

The summary embedding ${\bf{v}}_S(D)$ is expected to match the ``meaning'' of the reference summaries.
It should have the ability to properly judge the sentence saliency, which is consistent with the reference summaries.
Following \cite{kobayashi-noguchi-yatsuka:2015:EMNLP}, we use the cosine similarity between the summary embedding ${\bf{v}}_S(D)$ and a sentence embedding ${\mathbf{v}}(s)$ to predict the sentence saliency $r_s$.
\begin{equation} \label{EQ:similarity}
r_s= \frac{{{\bf{v}}(s) \bullet {\bf{v}}_S^T(D)}}{{||{\bf{v}}(s)|| \bullet ||{{\bf{v}}_S}(D)||}}
\end{equation}
That is why both document and summary embeddings are of the same dimensionality.

\subsection{Training}\label{SC:training}
We use the pre-trained word embeddings and do not update them to avoid over-fitting.
%To avoid over-fitting, we do not update word embeddings.
Thus, there are three types of weight matrices in our models, i.e., ${{\bf{W}}_\alpha }$, ${{\bf{W}}_\beta }$ and the transformation sub-matrices $({\bf{W}}_\gamma ^1, \cdots ,{\bf{W}}_\lambda ^{|C|})$.
Since the text classification dataset is much larger than the summarization dataset, ${{\bf{W}}_\alpha }$ and ${{\bf{W}}_\beta }$ are learned from the classification data only.
Yet, the transformation matrices have to be trained with the summarization data.
%Then, our summarization model focuses on the training of the set of the transformation matrices.

For text classification, we adopt the cross entropy as the cost function, i.e.,
\begin{equation}
{\varepsilon _C}(D) = \sum\nolimits_{i = 1}^{|C|} {1\{ c_D =  = i\} \ln {\bf{v}}_C^i(D)}, 
\end{equation}
where ${1\{ c_D =  = i\} }$ equals 1 iff the actual category is $i$.
Under this cost function, the gradient of softmax is similar to a linear function, which fastens the training process.

For summarization, we apply the pairwise ranking strategy~\cite{Collobert2011}  to tune the weights.
Specifically, each time we randomly select a sentence with a high actual saliency score and the other one with a low actual saliency score.
They are denoted as $s^+$ and $s^-$, respectively.
By Eq.~\ref{EQ:similarity}, we obtain their predicted saliency scores.
With the pairwise ranking criterion, TCSum should give $s^+$ a higher score in comparison with $s^-$.
Therefore the cost function is defined as follows:
\begin{equation} \label{eq:cost}
{\varepsilon _S}(D) = {\rm{ }}\max (0,\Omega  - r_{{s^ + }} + r_{{s^ - }}),
\end{equation}
where $\Omega$ is a margin threshold.

With the above two cost functions, we apply the diagonal variant of AdaGrad with mini-batches~\cite{duchi2011adaptive} to update model parameters.
AdaGrad adapts the learning rate for different parameters at different steps.
Thus it is less sensitive to initial parameters than the stochastic gradient descent.

\section{Experiments}
\subsection{Datasets} \label{SC:dataset}
\subsubsection{Summarization}
\begin{figure}
	\centering
	\includegraphics[width=0.7\linewidth]{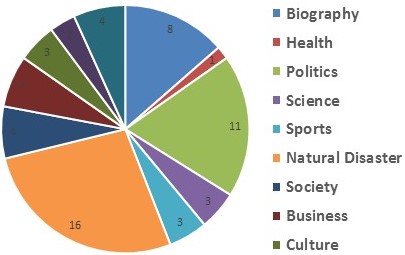}
	\caption{Category distribution on DUC 2002.}
	\label{fig:category}
\end{figure}

The most commonly used evaluation corpora for summarization are the ones published by the Document Understanding Conferences (DUC) and Text Analytics Conferences (TAC\footnote{\url{http://www.nist.gov/tac/} from 2007 $\sim$ now}).  
In this work, we focus on the generic multi-document summarization task, which was carried out in DUC 2001, 2002 and 2004. 
The documents are all from the news domain and a collection of documents related to the same topic are grouped together into a cluster.
Each cluster is accompanied by 2 to 4 reference summaries written by human experts.
Our summarization model compiles the documents in a cluster into \textit{a single document}.
Table \ref{TB:dataset} shows the size of the three datasets and the summary length limitation for each task. 
%Although DUC 2003 is query-focused, we use it as the extra training set.
%The preprocessing includes sentence splitting and tokenization, which is conducted by Stanford CoreNLP~\cite{manning-EtAl:2014:P14-5}.
The DUC datasets come from a wide range of categories, and we manually categorize the DUC documents into 11 categories, i.e., Biography, Culture, Business, Health, Politics, Law, Society, Natural Disaster, Science, Sports and International.
The category distribution of DUC 2002 is illustrated in Fig.~\ref{fig:category}.
Among these categories, Natural Disaster, Politics and Biography account for 60\% of the documents.

\begin{table}[ht]
	\centering
	\small
	\begin{tabular}{l|lllll}
		\hline
		Dataset & Cluster \# & Doc. \#  & Ref. \#  &Limitation\\ \hline
		DUC 01  & 30         & 309                & 60    & 100 words       \\ \hline
		DUC 02  & 59         & 567                 & 116   &100 words       \\ \hline
		DUC 04  & 50         & 500                & 200    &665 bytes      \\ \hline
	\end{tabular}
	\caption{Statistics of the summarization datasets.}
	\label{TB:dataset}
\end{table}

\subsubsection{Text Classification}
In order to benefit from text classification, we need to have a classification dataset large enough to cover all the 11 categories discovered in the DUC datasets.
We build such a dataset from the New York Times (NYT) Annotated Corpus \footnote{https://catalog.ldc.upenn.edu/LDC2008T19}.
The NYT corpus contains over 1.8 million articles published and annotated by the New York Times.
Notably, the New York Times is also an important data provider for DUC.
The NYT documents have rich metadata. 
We utilize three types of metadata (Types Of Material, Taxonomic Classifiers and Online Descriptors) to pick out the documents within those 11 categories.
%For example, the taxonomic classifier ``Top/News/Business'' indicates the corresponding document belongs to the category of Business, and the descriptor ``International Relations'' can filter documents from the category of International Relationship.
%We remove documents which belong to multiple categories.
We notice that the numbers of documents in different categories are extremely imbalanced.
For example, the category of Business contains more than 140,000 documents, while there are only 3,200 documents in the category of Natural Disaster.
Therefore, we conduct a sampling process to ensure that each category contains 3000-5000 documents. 
This classification dataset is about 30 times larger than the summarization dataset.

The cross validation shows that the learned classification model of TCSum achieves over 85\% accuracy on this dataset.
Since classification is not the focus of this paper, here we ignore the detailed performance evaluation of our classification model.

%\begin{table}[t]
%	\centering
%	\footnotesize
%	\begin{tabularx}{\linewidth}{X|X|X}
%		\hline
%		\textbf{Metadata}                   & \textbf{Value}                   & \textbf{Category}               \\ \hline
%		Types Of Material          & Biography               & Biography              \\ \hline
%		Taxonomic Classifiers      & Top/News/Sports         & Sports                 \\ \hline
%		Online Descriptors & International Relations & International \\ \hline
%	\end{tabularx}
%	\caption{Examples of metadata used to indicate categories.}
%	\label{tb:metadata}
%\end{table}
%\begin{figure}
%	\centering
%	\includegraphics[width=0.7\linewidth]{pic/CategoryNumber}
%	\caption{The category distribution of DUC 2002.}
%	\label{fig:CategoryNumber}
%\end{figure}

\subsection{Evaluation Metric for Summarization}
For evaluation, we use ROUGE\footnote{ROUGE-1.5.5 with options: -n 2 -m -u -c 95 -x -r 1000 -f A -p 0.5 -t 0. The parameter of length constraint is ``-l 100'' for DUC 2001/2002, and ``-b 665'' for DUC 2004.}~\cite{lin2004rouge}, which has been regarded as a standard automatic evaluation metric since 2004.
ROUGE measures summary quality by counting overlapping units such as N-grams, word sequences and word pairs between the candidate summary and the reference summary. 
Following the common practice, we take ROUGE-1 and ROUGE-2 recall scores as the main metrics for comparison.
ROUGE-1 and ROUGE-2 measure the uni-gram and bi-gram similarities, respectively.
%For example, the recall score of ROUEG-2 is computed as follows:
%\begin{equation}
%ROUGE - {2_{{\text{recall}}}} = \frac{{\sum\limits_{b \in \{ References\} } {{N_{match}}(b)} }}{{\sum\limits_{b \in \{ References\} } {N(b)} }}
%\end{equation}
%where $b$ stands for a bi-gram, and ${N_{match}}(b)$ is the maximum number of bi-grams co-occurring in a candidate summary and a set of reference summaries. $N(b)$ is the total number of bi-grams in the reference summaries.
During training, the actual saliency of a sentence (Eq.~\ref{eq:cost}) is also evaluated by ROUGE-2. 

\subsection{Model Settings}
For CNN, we introduce a word embedding set trained on a large English news corpus ($10^{10}$ tokens) using word2vec~\cite{mikolov2013efficient}.
The dimension of word embeddings is set to 50, as in many previous papers (e.g.,  \cite{Collobert2011}).
We also set the dimension of sentence and document embeddings equivalent the dimension of word embeddings, and the window size $h$ to 2, to be consistent with ROUGE-2 evaluation.
We empirically set the margin threshold of pairwise ranking $\Omega=0.1$. 
The initial learning rate is 0.1 and batch size is 128. 

A summary is obliged to offer both informative and non-redundant content.
While TCSum focuses on sentence ranking, it employs a simple greedy algorithm, similar to our previous work~\cite{cao2016attsum}, to select summary sentences.

\subsection{Baseline Methods} \label{SC:baseline}
%Two sentence regression baselines are designed. 
We compare TCSum with the best peer systems participating DUC evaluations, which are named as ``\textbf{Peer}'' plus their IDs.
%Afterwards, Support Vector Regression is used to learn the weights of features.
In addition, we include \textbf{R2N2}~\cite{cao2015ranking}\footnote{Although R2N2 can use integer linear programming to select better sentences, here we just consider the result of greedy selection for a fair comparison.}, a state-of-the-art supervised summarization model based on neural networks.
It applies the recursive neural network to learn the combination of hand-crafted features.
Notably, R2N2 still heavily depends on hand-crafted features.
By contrast, TCSum is fully data-driven, i.e., features are all learned automatically. 

We implement a widely-used learning-based summarization method Support Vector Regression  (\textbf{SVR})~\cite{li2007multi}.
It extracts a number of manually-compiled features from a sentence, such as TF (the frequency of a word in the cluster), CF(the number of documents containing this word in the cluster) and NUMBER (whether the sentence contains a number), etc. 
We also design three neural network based baselines, named as \textbf{NoTC}, \textbf{SingleT} and \textbf{EmSim}.
The first two are used to verify the value of text classification
NoTC does not use any classification data and just applies the summarization model of TCSum.
It is designed to check whether the summarization model can work alone.
SingleT ignores the predicted text category and uses a single transformation matrix.    
It explores the effect of summary styles.
The last one, EmSim, aims to test whether or not we need to learn the summary embedding. 
It just uses the cosine similarity between a sentence embedding ${\mathbf{v}}(s)$ and the document embedding ${\mathbf{v}}(D)$ to rank sentences.
EmSim is an unsupervised summarization model and similar to \cite{kobayashi-noguchi-yatsuka:2015:EMNLP}.
All these baselines employ the same sentence selection process as our model.
 
\subsection{Summarization Performance}
We conduct three-fold validation. 
The model is trained on two years' data and tested on the remaining year's.
The ROUGE scores of the models being compared are presented in Table~\ref{tb:rouge}.
We draw lines in this table to distinguish the models with and without hand-crafted features.

As can be seen, among the models completely dependent on automatically learned features, TCSum achieves highest performance on all the three datasets.
%We also find that TCSum has tighter confidential intervals than SingleT.
%That means the category-sensitive transformation indeed works.
The poor performance of EmSim denotes that we could not directly use the document embeddings learned from text classification to measure the sentence saliency for summarization.
Note that even NoTC achieves competitive performance with SVR.
Thus summarization models without hand-crafted features are doable. 
Meanwhile, SingleT greatly outperforms NoTC.
It verifies that text classification can indeed help a summarization model to learn better document representations.
Although TCSum does not always greatly surpass SingleT in terms of ROUGEs, we will show in the next section that it usually captures different summary styles.

Compared with other models, TCSum largely outperforms SVR and peer systems most of the time, and it is always superior to the state-of-the-art method R2N2.
Considering TCSum is not supplemented with any hand-crafted features, its performance is very promising.
After taking a closer look at the feature weights learned by SVR, we find the most important feature to measure sentence saliency is CF.
Since we treat the documents in a topic cluster as a single document, this feature is lost in our current summarization model.
It may be an important aspect that impedes the more excellent performance of TCSum.  

\begin{table}[]
	\centering
	\small
	\begin{tabular}{l|lll}
		\hline
		Year                  & Model   & ROUGE-1 & ROUGE-2 \\ \hline
		\multirow{7}{*}{2001} & Peer T  & 33.03   & \textbf{7.86}    \\
		& SVR     & 29.78   & 6.01    \\
		& R2N2    & 35.88   & 7.64    \\ \cline{2-4} 
		& NoTC    & 33.45   & 6.07    \\
		& EmSim   & 24.66   & 2.67    \\
		& SingleT & 35.22   & 7.42    \\
		& TCSum   & \textbf{36.45}   & 7.66    \\ \hline
		\multirow{7}{*}{2002} & Peer 26 & 35.15   & 7.64    \\
		& SVR     & 31.56   & 6.78    \\
		& R2N2    & 36.84   & 8.52    \\ \cline{2-4} 
		& NoTC    & 34.02   & 7.39    \\
		& EmSim   & 29.46   & 5.28    \\
		& SingleT & 36.54   & 8.44    \\
		& TCSum   & \textbf{36.90}   & \textbf{8.61}    \\ \hline
		\multirow{7}{*}{2004} & Peer 65 & 37.88   & 9.18    \\
		& SVR     & 36.18   & 9.34    \\
		& R2N2    & 38.16   & 9.52    \\ \cline{2-4} 
		& NoTC    & 35.66   & 8.66    \\
		& EmSim   & 30.80   & 5.07    \\
		& SingleT & 37.94   & 9.46    \\
		& TCSum   & \textbf{38.27}   & \textbf{9.66}    \\ \hline
	\end{tabular}
		\caption{ROUGE scores (\%) of different methods.}
		\label{tb:rouge}
\end{table}

\subsection{Discussion on Summary Style Learning}
% \usepackage{multirow}
\begin{table*}[t]
	\centering
	\small
	\begin{tabular}{l|l|p{11cm}}
		\hline
		Cluster               & Category         & Sentence                                                                                                                                                                   \\ \hline
		\multirow{3}{*}{D097} & Natural Disaster & The \textbf{storm}, packing \textbf{winds of up to 135 mph}, raged into \textbf{Charleston} \textbf{Thursday night}.                                                                                           \\ \cline{2-3} 
		& Biography        & ``This is a dangerous, \textbf{killer} hurricane, the likes of which few people who have lived all their lives in Charleston have experienced,'' warned \textbf{Mayor} Joseph P. Riley Jr.   \\ \cline{2-3} 
		& Politics         & \textbf{Gov.} Joe Frank Harris \textbf{declared} a \textbf{state of emergency} in six counties.                                                                                                       \\ \hline
		\multirow{3}{*}{D066} & Biography        & Sam Walton, \textbf{founder} of the Wal-Mart chain of discount supermarkets who died of cancer in April, negotiated these pitfalls much \textbf{better} than most.                           \\ \cline{2-3} 
		& Natural Disaster & By \textbf{1991} the chain's sales had risen to nearly \textbf{Dollars 44bn}, making it the world's largest retailer in terms of revenues, and the Walton family probably \textbf{America}'s richest. \\ \cline{2-3} 
		& Politics         & Bud is a senior \textbf{vice president} and board \textbf{member} of Wal-Mart.                                                                                                               \\ \hline
		\multirow{3}{*}{D076} & Politics         & Flamboyant former \textbf{Defense Minister} Hazeltine's challenge to \textbf{Prime Minister} Margaret Thatcher for \textbf{leadership} of the \textbf{Conservative Party} has caused a \textbf{political} sensation in Britain.   \\ \cline{2-3} 
		& Biography        & In the Persian Gulf crisis, she \textbf{boldly} joined with George Bush in sending troops to the Middle East.                                                                       \\ \cline{2-3} 
		& Natural Disaster & Among Western allies, she was alone at Ronald Reagan's side in \textbf{1986} in supporting the U.S. \textbf{bombing} of \textbf{Libya}.                                                               \\ \hline
	\end{tabular}
	\caption{Salient sentences selected by different categories. Sentences in the correct categories are displayed first.}
	\label{tb:summaryWithCategory}
\end{table*}

\begin{figure}
	\centering
	\includegraphics[width=0.9\linewidth]{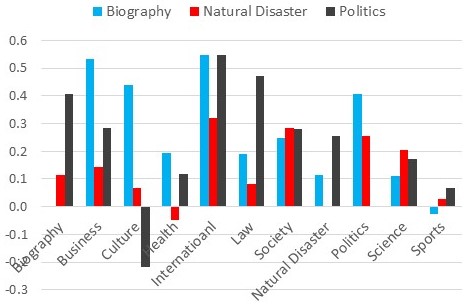}
	\caption{Similarity among the transformation matrices (we set the self similarity scores to 0).}
	\label{fig:CategoryMatrixSimilarity}
\end{figure}

We examine the ability of TCSum to learn summary styles in two ways.
At first, we speculate that similar transformation matrices tend to generate summaries with similar styles.
Therefore, we calculate the similarity among the transformation matrices $({\bf{W}}_\gamma ^1, \cdots ,{\bf{W}}_\lambda ^{|C|})$.
Here we flatten each matrix into a vector and use the cosine similarity to measure the similarity.
The scores of different transformation matrices are presented in Fig.~\ref{fig:CategoryMatrixSimilarity}. 
For ease of reference, we only show the results of three common categories on DUCs, i.e., Biography, Politics and Natural Disaster.
As can be seen, the similarity relations of these three categories vary greatly, which matches the intuition that the large difference of the summary styles exists among these categories.
For Biography, we find its transformation matrix is similar to 4 categories'. They are Business, Culture, Politics and International Relation.
One possible reason is that summaries in Biography necessarily tell the career-related information of a person.
Since DUC prefers choosing biographies about artists, businessmen and politicians, it is reasonable the summary style for Biography to be associated with these categories.
By contrast, Natural Disaster does not present obvious similarity to any other category.
We observe that summaries in Natural Disaster often contain a series of times, sites and numbers, while other categories seldom need so many details.
For Politics, we find it is similar to International Relationship and Law.
The former is understandable since we may use a number of terms of politics when describing international relationships.
The latter may be caused by the news content.
Many documents in this category are concerned with political scandals which often lead to lawsuits.
Interestingly, there is an obvious negative similarity between Politics and Culture.
The wordings in Politics are often thought to be serious while the documents in Culture are usually related to entertainment.

We also inspect the style change of the summaries generated according to different categories.
To this end, we manually assign a category to a document cluster and then calculate the sentence saliency based on our summarization model.
The salient sentences with respect to different categories are shown in Table~\ref{tb:summaryWithCategory}.
Due to the limit of space, we only display the top ranked summary sentences with the styles of three common text categories.
\begin{itemize}
	\item ``D097'' is about a hurricane (Natural Disaster).
	\item ``D066'' introduces the founder of Wall-Mart (Biography).
	\item ``D076'' describes the resignation of a prime minister (Politics).
\end{itemize}
As can be seen, the salient sentences calculated by the correct categories can properly represent the main idea of the document cluster.
Although ``D097'' and ``D066'' are not related to Politics, sentences selected by the corresponding transformation matrix still contain many terms of politics.
It is also shown that the three Biography sentences contain either the words describing the careers (killer, mayor, founder) or the evaluative words (better, boldly).
The career is a part of personal profile, and the description of main contributions of a person usually involves the evaluative words.
Therefore, the corresponding transformation matrix seems to well catch the two types of needs for Biography summaries.
We read the documents in ``D066'' and ``D076'' carefully, and find there is no sentence exactly matching Natural Disaster.
Thus it is not surprising that the sentences selected by Natural Disaster in these two clusters are somewhat strange.
However, we can see both sentences contain the date and site information.
This is absolutely consistent with the style that a summary of Natural Disaster is expected to have.
Moreover, both the money value and the word ``bombing'' can be used to describe the loss of a disaster.
It appears that, the transformation matrix for Natural Disaster still works well even on a topic other than Natural Disaster, with ``due diligence'' to complete its own task.

\section{Related Work}
Work on extractive summarization spans a large range of approaches.
Starting from unsupervised methods, one of the widely known approaches is Maximum Marginal Relevance (MMR) \cite{carbonell1998use}. It used a greedy approach to select sentences and considered the trade-off between saliency and redundancy.  
Good results could be achieved by reformulating it as an Integer Linear Programming (ILP) problem which was able to find the global optimal solution~\cite{mcdonald2007study,gillick2009scalable}.
Graph-based models such as Manifold~\cite{wan2009graph} played an important role in extractive summarization because of its ability to reflect various sentence relationships.
%For example, \cite{wan2009graph} adopted manifold ranking to make use of the within-document sentence relationships, the cross-document sentence relationships and the sentence-to-query relationships.
In contrast to these unsupervised methods, there are also many successful learning-based summarization approaches.
Different classifiers have been explored, including Conditional Random Field~\cite{galley2006skip}, Support Vector Regression~\cite{li2007multi} and Logistic Regression~\cite{li2013using}, etc.

Recently, the application of deep neural network techniques has attracted more and more interest in the summarization research. 
\cite{genest2011deep} used unsupervised auto-encoders to represent both manual and system summaries for summary evaluation.
%Their model generated representations of both manual and system summaries. 
%Then, the difference of these two representations were defined as features for regression, with regard to the Pyramid scores~\cite{passonneau2005applying}. 
Their method , however, did not surpass ROUGE.
%Similarly, \cite{SR2015} modeled summarization based on data reconstruction. 
%%They treat a summary as a compressed representation of documents. 
%Using a term frequency vector to represent a sentence, they aimed to find the proper subset which could reproduce the input documents.
%However, their system failed to achieve satisfactory performance.
%Later, \cite{yao2015compressive} improved this idea by introducing an additional sentence dissimilarity term in the optimization framework.
%However, the performance of their summary system is greatly inferior to state-of-the-art approaches like~\cite{wan2014ctsum}.
~\cite{cao2015ranking,cao2015learning} tried to use neural networks to complement sentence ranking features.
%\cite{cao2015ranking} applied recursive neural networks to automatically learn feature combination, while \cite{cao2015learning} learned document-independent features with convolutional neural networks.
Although the models achieved the state-of-the-art performance, they still relied on hand-crafted features.
A few researches explored to directly measure similarity based on distributed representations.
\cite{yin2015optimizing} trained a language model based on convolutional neural networks to project sentences onto distributed representations.
\cite{cheng2016neural} treated single document summarization as a sequence labeling task and modeled it by recurrent neural networks.
Others like~\cite{kobayashi-noguchi-yatsuka:2015:EMNLP} simply used the sum of trained word embeddings to represent sentences or documents.
In addition to extractive summarization, deep learning technologies have also been applied to compressive and abstractive summarization \cite{filippova-EtAl:2015:EMNLP,rush-chopra-weston:2015:EMNLP}.
%\cite{filippova-EtAl:2015:EMNLP} used word embeddings and Long Short Term Memory models (LSTMs) to output readable and informative sentence compressions.
%\cite{rush-chopra-weston:2015:EMNLP,hu-chen-zhu:2015:EMNLP} leveraged the neural attention model~\cite{bahdanau2014neural} in the machine translation area to generate one-sentence summaries.

\section{Conclusion and Future Work}
In this paper, we propose a novel summarization system called TCSum, which leverages text classification to improve the performance of summarization.
%It projects documents onto distributed representations shared by two tasks.
%Meanwhile, it presents summaries in various ways according to the predicted categories of documents.
Extensive experiments on DUC generic summarization benchmark datasets show that TCSum achieves the state-of-the-art performance, even without using any hand-crafted features.
We also observe that TCSum indeed catches the variations of summary styles among different text categories.
We believe our model can be used to other summarization tasks including query-focused summarization and guided summarization. 
In addition, we plan to let the model distinguish documents in a topic cluster, which is better adapted to the multi-document summarization.
%Moreover, with the summary embeddings in hand, we believe our system can be easily extended to abstractive summarization.
%The only additional step is to integrate a neural language model.

\section{ Acknowledgments}
The work described in this paper was supported by Research Grants Council of Hong Kong (PolyU 152094/14E), National Natural Science Foundation of China (61272291, 61672445) and The Hong Kong Polytechnic University (G-YBP6, 4-BCB5, B-Q46C).
The correspondence authors of this paper are Wenjie Li and Sujian Li.

\bibliographystyle{aaai}
\bibliography{ijcai2016}

\end{document}